%% file: main.tex
\documentclass[sigconf,authorversion,dvipsnames]{acmart}

\input{math_commands.tex}

\usepackage{multirow}
\usepackage[linesnumbered,algoruled,boxed,lined]{algorithm2e}
\usepackage{hyperref}
\usepackage{url}
\usepackage{pifont}

\usepackage{tikz}
\usetikzlibrary{decorations.text}
\usetikzlibrary{arrows.meta,bending,decorations.markings}
\usepackage{xcolor}
\newtheorem{remark}{Remark}

\newcounter{bxincomm}
\definecolor{aqua}{rgb}{0.00,0.67,0.80}

\newcounter{ygcounter}

\newcommand{\ygc}[1]{\ygc{\stepcounter{ygcounter}{\bf [YG's comment \arabic{ygcounter}: #1]}\;}}

\AtBeginDocument{%
  \providecommand\BibTeX{{%
    \normalfont B\kern-0.5em{\scshape i\kern-0.25em b}\kern-0.8em\TeX}}}

\begin{document}

\title{Spectral Transform Forms Scalable Transformer}

\author{Bingxin Zhou}
\authornote{Both authors contributed equally to this research.}
\email{bzho3923@uni.sydney.edu.au}
\orcid{0000-0002-3897-9766}
\affiliation{%
  \institution{The University of Sydney Business School, The University of Sydney}
  \city{}
  \country{Australia}}
\affiliation{%
  \institution{Institute of Natural Sciences, School of Mathematical Sciences,
  Shanghai Jiao Tong University}
  \city{}
  \country{China}
}

\author{Xinliang Liu}
\authornotemark[1]
\authornote{Correspondence author.}
\email{liuxinliang@sjtu.edu.cn}
\affiliation{%
  \institution{Institute of Natural Sciences, School of Mathematical Sciences,
  Shanghai Jiao Tong University}
  \city{}
  \country{China}
}

\author{Yuehua Liu}
\email{liuyh1214@gmail.com}
\affiliation{%
  \institution{Institute of Natural Sciences, School of Mathematical Sciences,
  Shanghai Jiao Tong University}
  \city{}
  \country{China}
}

\author{Yunying Huang}
\email{yunying.huang@sydney.edu.au}
\affiliation{%
  \institution{The University of Sydney Business School, The University of Sydney}
  \city{}
  \country{Australia}
}

\author{Pietro Li\`{o}}
\email{pl219@cam.ac.uk}
\affiliation{%
  \institution{Department of Computer Science and Technology, University of Cambridge}
  \city{}
  \country{United Kingdom}
}

\author{Yu Guang Wang}
\orcid{0000-0002-7450-0273}
\email{yuguang.wang@sjtu.edu.cn}
\affiliation{%
  \institution{Institute of Natural Sciences, School of Mathematical Sciences,
  Shanghai Jiao Tong University}
  \city{}
  \country{China}}
\affiliation{%
  \institution{School of Mathematics and Statistics, The University of New South Wales}
  \city{}
  \country{Australia}}

\begin{abstract}
Many real-world relational systems, such as social networks and biological systems, contain dynamic interactions. When learning dynamic graph representation, it is essential to employ sequential temporal information and geometric structure. Mainstream work achieves topological embedding via message passing networks (e.g., GCN, GAT). The temporal evolution, on the other hand, is conventionally expressed via memory units (e.g., LSTM or GRU) that possess convenient information filtration in a gate mechanism. Though, such a design prevents large-scale input sequence due to the over-complicated encoding. This work learns from the philosophy of self-attention and proposes an efficient spectral-based neural unit that employs informative long-range temporal interaction. The developed spectral window unit (\textsc{SWINIT}) model predicts scalable dynamic graphs with assured efficiency. The architecture is assembled with a few simple effective computational blocks that constitute randomized SVD, MLP, and graph Framelet convolution. The SVD plus MLP module encodes the long-short-term feature evolution of the dynamic graph events. A fast framelet graph transform in the framelet convolution embeds the structural dynamics. Both strategies enhance the model ability on scalable analysis. In particular, the iterative SVD approximation shrinks the computational complexity of attention to $\gO(Nd\log(d))$ for the dynamic graph with $N$ edges and $d$ edge features, and the multiscale transform of framelet convolution allows sufficient scalability in the network training. Our \textsc{SWINIT} achieves state-of-the-art performance on a variety of online continuous-time dynamic graph learning tasks, while compared to baseline methods, the number of its learnable parameters reduces by up to seven times.
\end{abstract}



\maketitle


\section{Introduction}
\label{sec:intro}
\begin{figure*}
    \centering
    \includegraphics[width=\textwidth]{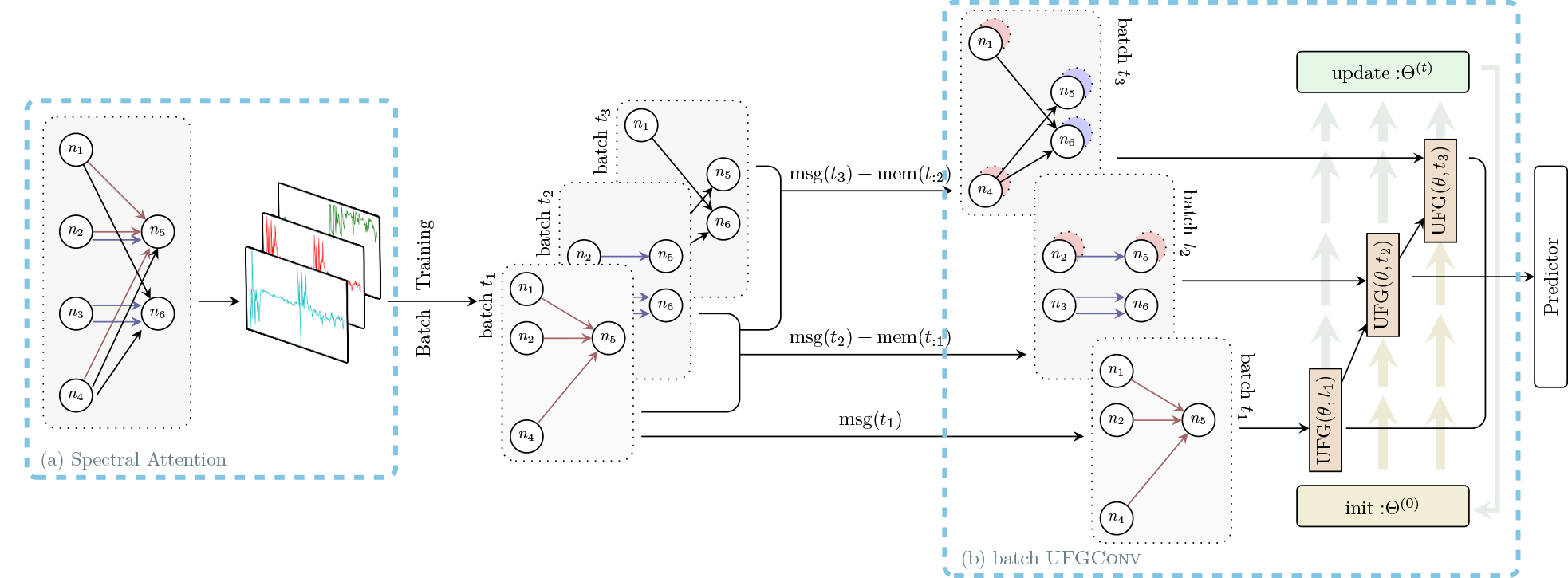}
    \caption{Architecture of the proposed Spectral Window Unit. (a) The input raw data is processed by a spectral attention module that extracts stable principal patterns of the feature dimension while encodes long-range time-dependency. (b) Local \textsc{UFGConv} embeds topological information of sliced subgraphs to node representations, which are sent to predictors for forecasting tasks.}
    \label{fig:architecture}
\end{figure*}
Dynamic graphs appear in many scenarios, such as transportation \cite{li2018diffusion}, pandemic spread \cite{panagopoulos2020transfer}, climate change \cite{cachay2021graph}, social network \cite{rossi2020temporal}, complex physics system \cite{sanchezgonzalez2020learning}, biology \cite{Gainza606202}. Learning dynamic graphs, however, is a challenging task where node features and graph structures evolve over time. The graph property requires being dynamically predicted, which needs a model to capture both time-dependent features and time-varying structures of a graph.

There are a few pioneering works on dynamic graph representation learning \citep{rossi2020temporal,xu2020inductive,kumar2019predicting}. Existing models usually embed sequential graph topology then feed into recurrent networks, similar to predicting on conventional time-series data. While the idea is intuitive and easy to follow, such a design can hardly generalize to continuous graphs. The embedding step happens on graph slices at each point of timer shift,  where node-by-node long-short term memory is impossible to preserve. Instead, the temporal graph network (\textsc{TGN}) \cite{rossi2020temporal} proposes to leverage a memory module that encodes previous records to the latest events. The event states are concatenated for describing node dynamics within a small time window. Nevertheless, this module has trouble in viewing the holistic graph evolving. The memory unit, e.g., gated recurrent unit (GRU), that the model relies on allows long-term interaction in a most implicitly way. Within the black box, there is no way to control or investigate the preserved message. Alternatively, self-attention \citep{vaswani2017attention} is a classic tool that improves the long-range memory for sequential data. In particular, it learns pair-wise similarity scores of the entire range of interest. This method preserves maximum dependency within a large window so that the memory is no longer a problem. However, for the cost of comprehensiveness, its complexity easily explodes when the timestamp grows rapidly.

This work proposes an effortless mechanism, where we show randomized SVD plus MLP play a similar role as a transformer \cite{dosovitskiy2020vit,vaswani2017attention} for dynamic data. SVD finds the pivotal parts of the temporal node information by multiplying a higher-order power of the feature matrix, which is analogous to the classical attention where the principal feature is learned by a linear neural layer.  In correspondence, SVD scales temporal features by weighting different components in its decomposition. The main patterns of transformed features are highlighted with a similarity measure from both the time dimension and node feature dimension. We thus call this procedure a \emph{spectral attention} mechanism. Moreover, the framelet graph convolution \citep{zheng2021framelets} provides sufficient scalability via its multilevel representation of the structured data and characterizes the non-Euclidean geometry of the graph data in evolutionary learning.

Both the spectral attention and framelet convolution are implemented in the fast algorithms with a much-reduced number of model parameters as compared to the classical schemes. The spectral attention and framelet convolution define a new computational unit for distilling information from dynamic graph data, which we call \emph{Spectral Window Unit}. It takes account of the three dimensions of time, node and structure features simultaneously, and it exploits the long-term evolution behavior of time-varying data. The fusion effect of three distinct dimensions that enables the effective extraction of temporal information is stimulated in the spectral domain under the transforms of SVD and framelets.

The rest of the paper is arranged as follows. Section~\ref{sec:relatedWork} reviews previous works that are closely related to our work. Section~\ref{sec:attnSVD} and Section~\ref{sec:GNN} presents the key components for temporal feature and structure embedding. In particular, we revisit the inefficient attention mechanism and identify an alternative spectral transform based on randomized SVD for potential improvement. Section~\ref{sec:model} details our proposed framework for dynamic graph processing. The model's empirical performance is reported in Section~\ref{sec:experiment} on inductive and transductive link prediction tasks, following extensive ablation study and computational efficiency analysis.


\section{Related work}
\label{sec:relatedWork}
This section reviews previous research that is mostly relevant to our work, which are graph representation learning and dynamic inference of sequential data.
    
\subsection{Graph Structure Embedding}
GNNs have seen a surge in interest and popularity recently. It has also shown great success in dealing with irregular graph-structured data that traditional deep learning methods such as CNNs fail to manage. The main factor that contributes to its success is that GNNs learn the structure pattern of graphs while CNNs can only handle regular grid-like inputs. Common to most GNNs and their variants
is the graph embedding through the aggregation of neighbor nodes (no matter in vertex domain or spectral domain after a certain transform) in a way of message passing \citep{gilmer2017neural,bronstein2017geometric,wu2020comprehensive}. Graph convolution is a key ingredient for graph embedding for node aggregation as similar to the convolution of pixels in CNNs. Convolution operating on vertex or nodes \citep{hamilton2017inductive, xu2018powerful} and convolutions on the pseudo-coordinate system that are mapped from nodes through a transformation (typically Fourier) \citep{bruna2013spectral}, correspond to spatial-based methods and spectral-based methods, respectively. 
Due to its intuitive characteristics of spatial-based methods which can directly generalize the CNNs to graph data with convolution on neighbors, most GNNs fall into the category of spatial-based methods \citep{atwood2016diffusion,velivckovic2017graph,monti2017geometric,huang2018adaptive,zhang2018end,li2018deeper,ying2018hierarchical,liu2019geniepath,xu2018powerful}. 

Many other spatial methods broadly follow the message passing scheme with different neighborhood aggregation strategies, but they are developed empirically being limiting theoretical understanding. In contrast, spectral-based graph convolutions \citep{bruna2013spectral,henaff2015deep,defferrard2016convolutional,levie2018cayleynets,zhao2021wgcn,xu2018graph,li2020fast,zheng2020decimated,zheng2020mathnet,zheng2021framelets} convert the raw signal or features in the vertex domain into the frequency domain. Spectral-based methods have already been proved to have a solid mathematical foundation in graph signal processing \citep{shuman2013emerging}, and the vastly equipped multi-scale or multi-resolution views push them to a more scalable solution of graph embedding. Versatile Fourier\citep{defferrard2016convolutional,kipf2016semi,henaff2015deep}, wavelet transforms\citep{zhao2021wgcn} and framelets\citep{zheng2021framelets} have also shown their capabilities in graph representation learning. Of these transforms, Fourier transforms is particularly one of the most popular ones and the work in \citep{lee2021fnet} gave a detailed review of how Fourier Transform enhances neural networks. In addition, with fast transforms being available in computing strategy, a big concern related to efficiency is well resolved.

\subsection{Temporal Encoding of Dynamic Graphs}
Recurrent neural networks (RNNs) are considered exceptionally successful for sequential data modelling, such as text, video, and speech \citep{graves2012sequence,graves2013speech,shahroudnejad2021survey}. In particular, Long Short Term Memory (LSTM) \citep{hochreiter1997long} and Gated Recurrent Unit (GRU) \cite{cho2014properties} gains great popularity in application. Compared to Vanilla RNN, they leverage a gate system to extract memory information, so that memorizing long-range dependency of sequential data becomes possible. Though, it is difficult to interpret their internal behaviors due to the complex network architecture \citep{chung2014empirical}. Also, the sequential data encoding prevents recurrent networks from efficiently capturing temporal dependencies. Instead, the Transformer network \citep{vaswani2017attention} designs an encoder-decoder architecture with the self-attention mechanism, so as to allow parallel processing on sequential tokens. The self-attention mechanism have achieved state-of-the-art performance across all NLP tasks \citep{vaswani2017attention,kumar2016ask} and even some image tasks \citep{dosovitskiy2020vit,yang2016stacked}. 


For dynamic GNNs, it is critical to consolidate the features along the temporal dimension. Dynamic graphs consist of discrete and continuous two types according to whether they have the exact temporal information \citep{skardinga2021foundations}. Recent advances and success in static graphs encourage researchers and enable further exploration in the direction of dynamic graphs. Nevertheless, it is still not recently until several approaches \citep{nguyen2018continuous,li2018deep,goyal2018dyngem,trivedi2018representation} were proposed due to the challenges of modeling the temporal dynamics. In general, a dynamic graph neural network could be thought of as a combination of static GNNs and time series models which typically come in the form of an RNN \citep{seo2018structured,manessi2020dynamic,narayan2018learning}. The first DGNN was introduced by Seo et al. \citep{seo2018structured} as a discrete DGNN and Know-Evolve \citep{trivedi2017know} was the first continuous model. JODIE \citep{kumar2019predicting} employed a coupled RNN model to learn the embeddings of the user/item. The work in \citep{sankar2020dysat} learns the node representations through two joint self-attention along both dimensions of graph neighborhood and temporal dynamics. The work in \citep{pareja2020evolvegcn} was the first to use RNN to regulate the GCN model, which means to adapt the GCN model along the temporal dimension at every time step rather than feeding the node embeddings learned from GCNs into an RNN. TGAT \citep{xu2020inductive} is notable as the first to consider time-feature interactions. Then Rossi et al. \cite{rossi2020temporal} presented a more generic framework for any dynamic graphs represented as a sequence of time events with a memory module added in comparison to \citep{xu2020inductive} to enable the short-term memory enhancement.

\section{Temporal Message Encoding}
\label{sec:attnSVD}
This section discusses the efficiency issue of scalable sequential feature embedding. The self-attention mechanism is first analyzed with its inevitable cost, following the power method-based randomized SVD that compacts an efficient version of the transformer.

\subsection{Linear Self-Attention and its Inefficiency}
We start from a simple linear attention without softmax activation \cite{cao2021transformer}. The attention for a given signal $\mX\in\R^{n\times d}$ reads
\begin{align} \label{func:attn}
    &{\rm attn}(\mX) := (\mQ\mK^{\top}\mV)/n, \\
    \text{where } &\mQ:=\mX\mW_Q, \mK:=\mX\mW_K,\mV:=\mX\mW_V. \notag
\end{align}
The three square matrices $\mQ$ (query), $\mK$ (key) and $\mV$ (value) contain learned basis functions, and they have an identical size of $d\times d$. As $d$ is typically smaller than the sample size $n$, this helps save the learning cost of algorithms, i.e., the number of parameters to approximate. The local attention applies trainable non-batch-based normalization on them. Alternatively, efficient attention \citep{shen2021efficient} leverages layer normalization out of the attention function. For now, we neglect this normalization operator as well as the scaler $n$ for simplicity, but the same intrinsic idea holds for both scenarios. 

Rewrite (\ref{func:attn}) with respect to $\mX$, we have
\begin{align} \label{func:attn2}
    {\rm attn}(\mX) = \mX\mW_Q\mW_K^{\top}\mX^{\top}\mX\mW_V.
\end{align}
This simple powerful module facilitates many applications in complicated data processing and analysis domains. Though, the compromised computational efficiency remains unsolved. The propagation of $\mQ\mK^{\top}$ $\mX\mW_Q\mW_K^{\top}\mX^{\top}$ requires an expensive matrix multiplication at $n\times n$. The computational cost explodes easily when the sample size grows drastically. To tackle this issue, we first revisit (\ref{func:attn2}) from the view of matrix decomposition and approximation.

\subsection{Randomized Power scheme of SVD}
Singular value decomposition (SVD) forms a unitary transform on the raw data signal \citep{bertero2020introduction}. A matrix $\mX\in\R^{n\times d}$ ($n>d$) is factorized into $\mU\Sigma\mV^{\top}$, where $\mU\in\R^{n\times n}$ and $\mV\in\R^{d\times d}$ are two orthonormal bases that span the row and column space of $\mX$. The raw matrix $\mX$ can be projected to some spectral domain by $\mU$ or $\mV$. For example, $\mX\mV$ denotes the spectral coefficients of $\mX$ where the projected space takes feature aggregation into account. As $\mX\mV=\mU\Sigma$, and $\mU\Sigma$ is a scaled row-space orthonormal basis, $\mX\mV$ can also be interpreted as an orthonormal basis of the raw matrix $\mX$. Alternatively, \emph{truncated SVD} \citep{hansen1987truncatedsvd} cuts off small singular value and their corresponding singular vectors so that $\mU\in\R^{n\times d'}$ and $\mV\in\R^{d\times d'}$ ($d'<d$) with the $d-d'$ most sensitive parts to small changes being removed. 
 
The randomized power scheme \citep{halko2011finding} improves the efficiency of SVD. The orthonormal basis of $\mX$ is approximated efficiently by the randomized power method with an iterative QR decomposition via the Gram-Schmidt algorithm. Formally, the basis
\begin{align} \label{func:QRsvd}
    \mQ = \mA\mR^{-1} = \mX(\mX^{\top}\mX)^q\mR^{-1},
\end{align}
where $\mA$ denotes the power-$q$ approximation of $\mX$. The $\mR$ is an upper triangular matrix determined by column elements of $\mX$ and $\mQ$. For illustration purposes, we eliminate the random factor $\Omega$ that usually produces dimensionality reduction on $\mY$. Compared to vanilla SVD, $(\mX^{\top}\mX)^q\mR^{-1}$ serves as an approximator of the column singular vectors $\mV$. The closeness of the approximation to the exact solution is controlled by iteration $q$. The more iterations applied, the larger gaps are discovered between different basis vectors, and the smaller the approximation error is involved.

\subsection{Linear Attention versus Randomized SVD}
We now connect linear attention with randomized SVD. The current SVD approximation in (\ref{func:QRsvd}) involves no learnable scheme, thus it is impossible to adapt the importance of orthonormal basis (or spectral coefficient) with respect to the input data. One simple solution is endowing a linear transformation
\begin{align} \label{func:QRsvd2}
    \widetilde{\mQ} = \mQ\mW = \mX(\mX^{\top}\mX)^q\mW.
\end{align}
We unite $\mR^{-1}$ with the learnable parameter $\mW$ to abbreviate the representation. An intuitive explanation of this linear transformation is to rearrange and summarize the principal factors (singular vectors) to the importance of describing entity features. While the QR decomposition ranks the orthonormal basis according to their energy decreasing, this measure of importance might be less practical for many prediction tasks. Thus it is preferred to allow data-driven adjustments on principal components.

If we rewrite $\mW_Q\mW_K^{\top}=\mW_1$ and $\mW_V=\mW_2$ of (\ref{func:attn2}):
\begin{align} \label{func:attn3}
    {\rm attn}(\mX) = \mX\mW_1\mX^{\top}\mX\mW_2.
\end{align}
Both (\ref{func:QRsvd2}) and (\ref{func:attn3}) share a similar format except for the number of learnable parameter sets and the power scheme. In fact, the linear self-attention in (\ref{func:attn3}) can be considered as a special case of (\ref{func:QRsvd2}) in the sense that the attention mechanism implicitly calculates a $1$-iteration QR approximation of SVD basis. While the power-$1$ approximation is of limited accuracy, attention makes linear adjustments by learnable parameter that approximates $\mW_1\mX^{\top}\mX\mW_2 \approx \mX(\mX^{\top}\mX)^q\mR^{-1}\mW$. 

In fact, both $\mW_1\mX^{\top}\mX\mW_2$ (or $\mK^{\top}\mV$) and $\mX(\mX^{\top}\mX)^q\mR^{-1}\mW$ serve the same role as a similarity metric of $\mX$'s row space. The $\mX^{\top}\mX$ in a one-step approximation of QR iteration aggregates row-wise variation and summarize a low-rank covariance matrix of the feature space. The subsequent power iterations and adjustments on $\mR$ only widen the gap between different modes so that more resources can be focused on large modes, and the smallest modes (which are usually considered as noise or disturbance of truth) are eventually removed. The same procedure on attention, in contrast, is highly dependent on the input value. The learning process is indeed a black-box learning module. While the algorithm is considerably flexible and adaptive to a broad range of scenarios, such advantages are traded with precision and controlling power.

The other pain point of the attention mechanism in (\ref{func:attn3}) is the computational expense on large datasets. Consider an extremely long sequence of input $\mX_{N}\in\R^{N\times d}$ where $N \gg d$. The attention method have to calculate $\mX\mW_1\mX^{\top} \in \R^{N\times N}$ following another $N\times d$ matrix multiplication. This huge square matrix could result in a great burden on both calculation speed and storage. In comparison, (\ref{func:QRsvd2}) starts from $\mX^{\top}\mX\in\R^{d\times d}$ that controls the main calculation within an acceptable dimension. While the length of sequential data, such as time-series data, can easily be expanded, this `dimension reduction' trick is important for scalable learning tasks.


\section{Graph Topology Embedding}
\label{sec:GNN}
We now present the two modules with respect to sub-graph topology embedding. The first module is called a memory window that batchrizes the event flow with fixed batch size. The construction of the adjacency matrix relies on the events that happened within this batch, and the features constitute both present and (recent) previous information of the nodes. The second is a graph convolution for topology representation learning. We use a spectral-based method in consideration of scalability.

\subsection{Memory Window}
\label{sec:tgnWindow}
A continuous-time dynamic graph records the temporal evolution by a sequence of events. Instead of treating graph intervals as discrete slices, Temporal Graph Networks (\textsc{TGN}) \cite{rossi2020temporal} practices node embedding by a message-memory encoder. Given an event $e_{i}[t]$ at time $t$ with respect to node $n_i$, we name it a \emph{message} of $n_i$ at time $t$, denoted as ${\rm \mathbf{\mathbf{msg}}}(e_{i}[t])$. In addition, if the node was previously recorded active, we use ${\rm \mathbf{mem}}(e_{i}[:t])$ to represent the past information, or \emph{memory}, of $n_i$ prior to time $t$. The memory module ${\rm \mathbf{mem}}()$ refreshes constantly with the latest messages to capture the dynamic nature of graph interactions. When a new event $e_i[t]$ is recorded, the updated memory at time $t$ is
\begin{equation*}
    {\rm \mathbf{mem}}(e_{i}[t])=f({\rm \mathbf{mem}}(e_{i}[:t]), {\rm \mathbf{msg}}(e_{i}[t]))
\end{equation*}
with a trainable function $f()$. Depending on when the node $i$ was previously recorded, the last memory can be found before $t-1$. Also it is possible to recall memory from more than one step away. 

We thus describe $n_i[t]$'s state by its hidden memory $h_i[t]$ at time $t$, which concatenates ${\rm \mathbf{msg}}(e_{i}[t])$ and ${\rm \mathbf{mem}}(e_{i}[:t])$, i.e., 
\begin{equation} \label{func:msg_mem_encoder_node}
    h_i(t) = {\rm \mathbf{concat}}({\rm \mathbf{msg}}(e_{i}[t])) \| {\rm \mathbf{mem}}(e_{i}[:t]))).
\end{equation}
The embedding for an interactive event $e_{ij}(t)$ between two nodes $n_i$ and $n_j$ is similar, which reads
\begin{equation}\label{func:msg_mem_encoder_edge}
    h_i(t) = {\rm \mathbf{concat}}({\rm \mathbf{msg}}(e_{ij}[t]) \| {\rm \mathbf{mem}}(e_{i}[:t]) \| {\rm \mathbf{mem}}(e_{j}[:t])).
\end{equation}

\subsection{Framelet Graph Transforms}
\label{sec:UFGconv}
Graph convolution is a key ingredient for graph representation learning. Given an undirected graph $\gG=(\sV,\mathbb{E},\mX)$ with $N=|\sV|$ nodes, its edge connection is described by an adjacency matrix $\mA\in \R^{N\times N}$ and the $d$-dimensional node feature is stored in $\mX\in\R^{N\times d}$. Graph convolution aims at encoding $\mA,\mX$ to a hidden representation $\mH$ for prediction tasks. Depending on whether the convolution is conducted on the vertex domain or a transformed domain, typical methods are classified into either spatial or spectral methods. Here we are interested in spectral methods, or more specifically, Frame wavelet (Framelet) methods, where multilevel or multiscale feature allows scalable graph representation learning. 

We now brief the undecimated Framelet graph convolution (\textsc{UFGConv}) method proposed by \cite{zheng2021framelets}, which leverages fast Framelet decomposition and reconstruction for graph topology embedding. 

The Framelet convolution defines in a similar manner to any typical spectral graph convolution layer that
\begin{equation}
    \vtheta\star \mX = \gV {\rm diag}(\vtheta) \gW \mX',
\end{equation}
where $\mX'$ is transformed (hidden) feature embedding and $\vtheta$ denotes learnable parameters. The $\gW$ and $\gV$ denotes the decomposition and reconstruction operators that transform the graph signal between the vertex domain and the Framelet domain. The fast approximation of Framelet coefficients is crucial for an efficient \textsc{UFGConv} algorithm, and we now brief the graph Framelet transforms \citep{dong2017sparse,zheng2021framelets}.

Framelet transform divides an input signal to multiple channels by a set of low-pass and high-passes \emph{Framelet bases}. For a specific nodes $p$, its bases at \emph{scale level} $l=1,\ldots,L$ reads
\begin{align*}
    \boldsymbol{\varphi}_{l,p}(v) &= \sum_{\ell=1}^{N} \hat{\alpha}\left(\frac{\lambda_{\ell}}{2^{l}}\right)
    \overline{\vu_{\ell}(p)}\vu_{\ell}(v)\\
    \boldsymbol{\psi}_{l,p}^k(v) &= \sum_{\ell=1}^{N} \widehat{b^{(k)}}\left(\frac{\lambda_{\ell}}{2^{l}}\right)\overline{\vu_{\ell}(p)}\vu_{\ell}(v).
\end{align*}
We call $\Psi=\{\alpha;\beta^{(1)},\dots,\beta^{(K)}\}$ a set of \emph{scaling functions}, and it is determined by a filter bank $\mathbf{\eta}:=\{a;b^{(1)},\dots,b^{(K)}\}$. The other component is the eigen-pairs $\{(\lambda,\vu)\}_{j=1}^{N}$ of the graph Laplacian $\gL$, which plays a key role for embedding graph topology. The Framelet basis projects input signals to a transformed domain as \emph{Framelet coefficients}. Given a signal $\vx$, $\langle\boldsymbol{\varphi}_{l,p},\vx\rangle$ and $\langle\boldsymbol{\psi}_{l,p}^k,\vx\rangle$ are the corresponding Framelet coefficients for node $p$ at scale $l$. 

To allow fast approximation of the filter spectral functions, $m$-order Chebyshev polynomials is considered. We denote $m$-order approximation of $\alpha$ and $\{\beta^{(1)},\dots,\beta^{(K)}\}$ by $\gT^m_0$ and $\{\gT^m_{k}\}_{k=1}^{K}$. At a given level $L$, the full set of Framelet coefficients $\gW_{k,1}\vx$ reads
$$
\begin{cases}
    \gT^m_k(2^{-H}\gL)\vx, & l=1 \\
    {\gT^m_k}(2^{-H -l}\gL){\gT^m_0}(2^{-H-l+1}\gL){\cdots\gT^m_0}(2^{-H}\gL)\vx, & \text{otherwise},
\end{cases}
$$
where the dilation scale $H$ satisfies $\lambda_{\max} \leq 2^H\pi$. 


\section{Spectral Window Unit}
\label{sec:model}
This section presents the proposed Spectral Window Unit (\textsc{SWINIT}) model for dynamic graph representation learning. We divide the learning task into two steps. First, a \emph{spectral attention} module is leveraged for efficient temporal feature encoding. The new representations are then batchrized and sent to \textsc{UFGConv} for topological encoding. The particular designs are detailed below. Algorithm~\ref{alg:SWINIT} summarizes the complete training process with the model complexity analysis be provided. The model architecture is briefed in Figure~\ref{fig:architecture}. In addition, Figure~\ref{fig:SVD} and Figure~\ref{fig:UFG} give detailed demonstrations for each portion of the proposed model for better understanding.

\subsection{Temporal Feature Encoding}
\label{sec:svdEmbedding}
\begin{figure}
    \centering
    \input{tikz_SVD}
    \caption{A dissection of Randomized SVD-oriented spectral message encoder on $\mX$. Section~\ref{sec:svdEmbedding} discusses the two interpretations of this embedding in detail.}
    \label{fig:SVD}
\end{figure}
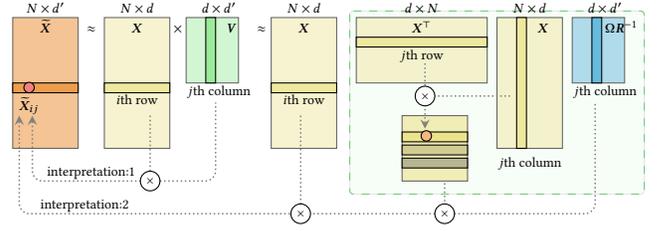
Our method embeds long-range time-dependency by a randomized SVD-based spectral attention mechanism, which implements a similar function to the classic attention design but provides additional scalability and reliability, as analyzed in Section~\ref{sec:attnSVD}. This section gives an intuitive justification for the feasibility of dynamic events processing by spectral attention.

Consider a raw matrix $\mX\in\R^{N\times d}$ that records sequential events of a period. Rather than directly training prediction models on $\mX$, it is desired to first find a low-dimension projection $\widetilde{\mX}$ to some spectral domain that. The $\widetilde{\mX}$ is believed better summarize the principal patterns of the input space and is immune to a minor disturbance. Consequently, they are easier to train with small chunks later on. To this end, we propose to encode the raw input with a randomized power scheme of truncated SVD. Similar to the self-attention mechanism, our spectral encoder reweights the input with their similarity score. Moreover, the similarity matrix is updated with explicit rules, so that the learning process is more manageable in comparison to a data-driven black-box approximation. We provide two interpretations to help understand how exactly this spectral attention method summarizes main patterns from both directions of feature and time. 

\paragraph{Interpretation 1} \textbf{The spectral attention extracts information of feature dimension by $\widetilde{\mX} \approx \mX\mV$}: We know from definition $\mX:=\mU\Sigma\mV^{\top}$ that SVD stores factorized features (columns) in $\mV$, and the factorized timestamps (rows) in $\mU$. The truncated SVD is design especially when the input is of low-rank, or full-rank but with noise. A purified input is transformed by $\widetilde{\mX}\approx \mX\mV\in\R^{N\times d'} (d>d')$. Consequently, the projected samples of $\widetilde{\mX}$ from $\mV$ finds a new representation by referring most effective feature dimension representations. For example, the $j$th feature of the $i$th transformed sample $\widetilde{\mX}_{ij}=\mX_i\mV_j$. The raw $\mX_i$ is concreted to a point following the projection rule of the $j$th factorized feature. Similar mappings by the $1, 2, \dots, d'$th factorization of $\mV$ converts the raw $\mX_i$ to a new space that is mostly representative from the perspective of features.

\paragraph{Interpretation 2} \textbf{The spectral attention aggregates information of time dimension by $\tilde{\mX} \approx \mX(\mX^{\top}\mX)\mR^{-1}$}: To get the full picture of the information aggregation, let's focus on the simplest case of iteration $q=1$. We already know from Section~\ref{sec:attnSVD} that the $\mX^{\top}\mX\mR^{-1}$ (in green box) is a one-step approximation of $\mV$. Here we show how this approximation embeds the long-range time-dependency of the input data. Figure~\ref{fig:SVD} show how the $j$th element of the $j$th row in $\mX^{\top}\mX$ is calculated by the inner product of the $j$th column vector of $\mX$. The calculated $d\times d$ matrix is indeed a covariance matrix that summarizes the column-wise linear relationship of $\mX$. In our case, the $j$th column of $\mX$ describes the evolution of the $j$th feature over the entire timeline, and the covariance matrix gives a linear similarity measure of all features based on their temporal trace. If we project the raw matrix $\mX$ by this similarity matrix, the new representation will include the temporal correlation of all time. The same interpretation holds after $q$ iterations. While the algorithm concentrates high energies on more expressive modes and takes some linear adjustments (via $\mR^{-1}$), the fundamental format of the covariance matrix $\mX^{\top}\mX$ is never changed.

\subsection{Structured Node Embedding}
\begin{figure}
    \centering
    \input{tikz_UFG}
    \caption{Inheritance and renewal of the global spectral coefficient $\Theta$. The first row indicates the alternation of the global $\Theta$ from $t-1$ to $t+1$. The second row showcases the local $\theta$ updates by \textsc{UFGConv} of a graph at step $t$ and $t+1$.}
    \label{fig:UFG}
\end{figure}
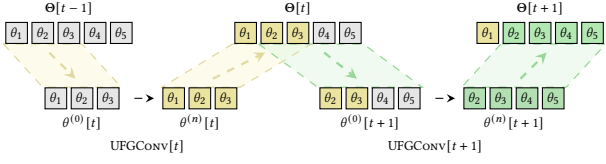
So far the event-based data have been projected to a spectral domain of the feature dimension. While the long-term dependencies are well-encoded during this process, the short-term memory requires further enhancement. In addition, the intrinsic structural information is waiting for embedding. To this end, a memory window gets involved to divide multiple batches of subgraphs. This operation allows zooming into a small window of events and extracting critical messages. The current state of an underlying node is also closely connected to recent messages of the same entity (node) from previous states. To include this part of the information, which we call memories, an aggregation is employed for assembling. The importance of short-term memorization has been verified by \cite{rossi2020temporal}. We adopt a similar idea for this slicing and aggregation by following the procedures introduced in Section~\ref{sec:tgnWindow}.

Now that the graph sequence is prepared, the next step is graph convolution for structure embedding. We employ \textsc{UFGConv}, a special type of spectral-based graph convolution, that allows potential scalability over multi-level graph signal processing. In classic graph-level representation learning tasks, \textsc{UFGConv} can easily parallel to multiple graphs that are independent of each other. For sequential graphs, the time-dependency is required to be implanted by a properly designed intra-connection on transform parameters $\theta$s. Inspired by the fact that the parameters are stored on a diagonal line and they scale the node-wise Framelet coefficients, we link the initialization of current Framelet coefficients with their best estimation of the last record, i.e., 
\begin{align*}
    \theta_i[t]^{(0)} = \theta_i[:t]^{(n)}
\end{align*}
for the $i$th $\theta$ value with respect to node $i$ at time $t$. Here $0$ denotes the first iteration (or initial value) of $\theta$ and $n$ is the estimated $\theta$ after $n$ propagation. The full set of $\theta$ is stored in a global $\Theta$, so that every time when step into a new \textsc{UFGConv}, the algorithm extracts previous $\theta$s from $\Theta$. The updated parameters of the training procedure, while being used for prediction, are also restored to $\Theta$ for next calling. Figure~\ref{fig:UFG} gives a simple demonstration. Consider a small graph with $5$ nodes. The global $\Theta$ has $5$ parameters to tune. The subgraph at batch $t$ contains the first $3$ of the $5$ nodes. When training the \textsc{UFGConv} layer, the model inherits the best estimation of $\{\theta_1,\theta_2,\theta_3\}$ before $t$ as initialization. The optimized model is deployed for further prediction tasks. Meanwhile, $\Theta$ replaces $\{\theta_1,\theta_2,\theta_3\}$ by the three updated parameters. This process is conducted repetitively along with the training procedure until all subgraphs are trained and evaluated.

\begin{algorithm}[t]
    \hsize=\textwidth
    \SetKwData{step}{Step}
    \SetKwInOut{Input}{Input}\SetKwInOut{Output}{Output}
    \BlankLine
    \Input{raw sequential data $\mX$}
    \Output{label prediction $\mY$}
    Initialization: global $\Theta$\\
    \textcolor{RoyalBlue}{\bf Randomized SVD $\mX\mV\leftarrow\mX(\mX^{\top}\mX)^q\mR^{-1}$};\\
    \For{batch $i \leftarrow 1$ \KwTo $N-2$}{
        \bf $h_i[t]\leftarrow {\rm \textbf{msg}}(e_{i}[t])\|{\rm \textbf{mem}}(e_{i}[:t])$;\\
        \textcolor{RoyalBlue}{$\gG_i\leftarrow(\mA_i,\mX_i\leftarrow\text{\textbf{FC}}(h_i[t])$)};\\
        \textcolor{RoyalBlue}{$H_i\leftarrow\textsc{\textbf{ UFGConv}}(\mA_i,\mX_i,\theta_i)$};\\
        $\mY_i\leftarrow\text{Predictor}(H_i)$;\\
        $\Theta_i\leftarrow\theta_i$;\\
        $\mY_{\text{i,val}}\leftarrow\text{Predictor}(H_{i+1})$;\\
        $\mY_{\text{i,test}}\leftarrow\text{Predictor}(H_{i+2})$;\\
        Update: score($\mY_{\text{val}}$), score($\mY_{\text{test}}$).
    }
    \caption{\textcolor{RoyalBlue}{\bf \textsc{SWINIT}: Spectral Window Unit}}
    \label{alg:SWINIT}
\end{algorithm}

\subsection{Complexity Analysis}
As mentioned in Section~\ref{sec:attnSVD}, our design of spectral attention is efficient with a small time and space complexity, as we analyze as follows.
We analyze the computational complexity for the SWINIT in Algorithm~\ref{alg:SWINIT} by estimating the cost for the three main computational units: randomized SVD for the entire training data, and MLP and UFGConv (framelet convolution) for batched graphs.  

\paragraph{Time complexity}  For a dynamic graph with $N$ events (edges) and $d$ edge features, the computational cost for randomized SVD is $\gO(Nd\log(d))$ \cite{halko2011finding}. The MLP has cost $\gO(N)$ in total. For all $m$ batches, the framelet convolution (UFGConv) has the complexity of $\gO(\sum_{i=1}^m N_i S_i \log_2(\lambda_i/\pi) F)$ where $N_i, S_i$ are the number of edges and sparsity of the $i$th batched graph, $\lambda_i$ is the largest eigenvalue of the corresponding graph Laplacian \cite{zheng2021framelets}, and $F$ is the number of the node features. In practice, the $N_i$ for each batched graph can be set as $N/m$, and we suppose $S_i$ and $\lambda_i$ are bounded by constants. The total computational cost of  \textsc{SWINIT} is $\gO\left(N(d\log(d)+F)\right)$.

\paragraph{Space complexity} For the randomized SVD, the memory cost is $\gO(Nd)$. The MLP with $l+1$ fully connected layers needs memory $\gO(n_1\sqrt{N/m}+n_1\times m_1+\cdots+n_l\times m_{l})$. Suppose each layer has the same number of hidden neurons $n$, then MLP has the space complexity $\gO(n\sqrt{N/m}+n^2l)$. The memory cost of framelet convolution is $\gO(NF)$. Then, the total space complexity of \textsc{SWINIT} is $\gO\left(N(d+F)+n\sqrt{N/m}+n^2l\right)$. 

\paragraph{Parameter number} The trainable network parameters appear mainly in MLP and UFGConv. As similar to space complexity analysis, \textsc{SWINIT} has the $\gO\left(n\sqrt{N/m}+n^2l+NF\right)$ parameters in total. 


\section{Numerical Examples}
\label{sec:experiment}

\begin{table*}[t]
    \caption{Performance of link prediction over 10 repetitions}
    \label{tab:linkPrediction}
    \begin{center}
    \resizebox{0.9\linewidth}{!}{
    \begin{tabular}{llrrrrrrr}
    \toprule
    && & \multicolumn{2}{c}{\textbf{Wikipedia}} & \multicolumn{2}{c}{\textbf{Reddit}} & \multicolumn{2}{c}{\textbf{MOOC}}\\ \cmidrule(lr){4-5}\cmidrule(lr){6-7}\cmidrule(lr){8-9}
    &\textbf{Model} & \# parameters & precision & ROC-AUC & precision & ROC-AUC & precision & ROC-AUC \\ 
    \midrule
    \multirow{5}{*}{\rotatebox[origin=c]{90}{transductive}} \hspace{2mm}
    &\textsc{DyRep}\citep{trivedi2019dyrep} & $920\times 10^{3}$ & $94.67${\scriptsize $\pm0.25$} & $94.26${\scriptsize $\pm0.24$} & $96.51${\scriptsize $\pm0.59$} & $96.64${\scriptsize $\pm0.48$} & $79.84${\scriptsize $\pm0.38$}& $81.92${\scriptsize $\pm0.21$} \\
    &\textsc{JODIE-rnn}\citep{kumar2019predicting} & \textcolor{violet}{\bm{$209\times 10^{3}$}} & $93.94${\scriptsize $\pm2.50$} & $94.44${\scriptsize $\pm1.42$} & $97.12${\scriptsize $\pm0.57$} & $97.59${\scriptsize $\pm0.27$} & $76.68${\scriptsize $\pm0.02$}&$81.40${\scriptsize $\pm0.02$} \\ 
    &\textsc{JODIE-gru}\citep{kumar2019predicting} & \bm{$324 \times 10^{3}$} & $96.38${\scriptsize $\pm0.50$} & \textcolor{violet}{\bm{$96.75${\scriptsize $\pm0.19$}}} & $96.84${\scriptsize $\pm0.39$} & $97.33${\scriptsize $\pm0.25$} &$80.29${\scriptsize $\pm0.09$}  & \bm{$84.88${\scriptsize $\pm0.30$}} \\ 
    &\textsc{TGN-gru}\citep{rossi2020temporal} & $1,217\times 10^{3}$ & \bm{$96.73${\scriptsize $\pm0.09$}} & $96.45${\scriptsize $\pm0.11$} &  \textcolor{violet}{\bm{$98.63${\scriptsize $\pm0.06$}}} & \textcolor{violet}{\bm{$98.61${\scriptsize $\pm0.03$}}} & \textcolor{violet}{\bm{$83.18${\scriptsize $\pm0.10$}}} & $83.20${\scriptsize $\pm0.35$}\\ 
    &\textsc{SWINIT-mlp} (ours) & \textcolor{red}{\bm{$170\times 10^{3}$}} & \textcolor{violet}{\bm{$97.02${\scriptsize $\pm0.06$}}} & \bm{$96.51${\scriptsize $\pm0.08$}} & \bm{$98.19${\scriptsize $\pm0.05$}} & \bm{$98.15${\scriptsize $\pm0.06$}}  & \bm{$82.40${\scriptsize $\pm0.24$}} &\textcolor{violet}{\bm{$85.55${\scriptsize $\pm0.17$}}}  \\  
    &\textsc{SWINIT-gru} (ours) &$376\times 10^{3}$&\textcolor{red}{\bm{$97.44${\scriptsize $\pm0.05$}}}  & \textcolor{red}{\bm{$97.15${\scriptsize $\pm0.06$}}}& \textcolor{red}{\bm{$98.69${\scriptsize $\pm0.09$}}}  & \textcolor{red}{\bm{$98.66${\scriptsize $\pm0.12$}}}&\textcolor{red}{\bm{$84.50${\scriptsize $\pm0.10$}}} &\textcolor{red}{\bm{$86.88${\scriptsize $\pm0.09$}}}\\ 
    \midrule
    \multirow{5}{*}{\rotatebox[origin=c]{90}{inductive}} \hspace{2mm}
    &\textsc{DyRep}\citep{trivedi2019dyrep} & $920\times 10^{3}$ & $92.09${\scriptsize $\pm0.28$} & $91.22${\scriptsize $\pm0.26$} & $96.07${\scriptsize $\pm0.34$} & $96.03${\scriptsize $\pm0.28$} & $79.64${\scriptsize $\pm0.12$}& $82.34${\scriptsize $\pm0.32$} \\
    &\textsc{JODIE-rnn}\citep{kumar2019predicting} &  \textcolor{violet}{\bm{$209\times 10^{3}$}}& $92.92${\scriptsize $\pm1.07$}  & $92.56${\scriptsize $\pm0.87$} & $93.94${\scriptsize $\pm1.53$} & $95.08${\scriptsize $\pm0.70$} & $77.17${\scriptsize $\pm0.02$}&$81.77${\scriptsize $\pm0.01$} \\ 
    &\textsc{JODIE-gru}\citep{kumar2019predicting} & $324 \times 10^{3}$& \textcolor{violet}{\bm{$94.93${\scriptsize $\pm0.15$}}} & \textcolor{violet}{\bm{$95.08${\scriptsize $\pm0.70$}}} & $92.90${\scriptsize $\pm0.03$} &$95.14${\scriptsize $\pm0.07$}&$77.82${\scriptsize $\pm0.17$}  & \bm{$82.90${\scriptsize $\pm0.60$}} \\ 
    &\textsc{TGN-gru}\citep{rossi2020temporal} & $1,217\times 10^{3}$& \bm{$94.37${\scriptsize $\pm0.23$}} & \bm{$93.83${\scriptsize $\pm0.27$}} & \textcolor{violet}{\bm{$97.38${\scriptsize $\pm0.07$}}} & \textcolor{violet}{\bm{$97.33${\scriptsize $\pm0.11$}}} & \bm{$81.75${\scriptsize $\pm0.24$}}& $82.83${\scriptsize $\pm0.18$}\\ 
    &\textsc{SWINIT-mlp} (ours) &  \textcolor{red}{\bm{$170\times 10^{3}$}} & $94.27${\scriptsize $\pm0.05$} & $93.28${\scriptsize $\pm0.05$} &  \textcolor{red}{\bm{$97.49${\scriptsize $\pm0.01$}}} & \textcolor{red}{\bm{$97.34${\scriptsize $\pm0.02$}}}  & \textcolor{red}{\bm{$82.54${\scriptsize $\pm0.08$}}}& \textcolor{red}{\bm{$85.23${\scriptsize $\pm0.09$}}}\\ 
    &\textsc{SWINIT-gru} (ours) & \bm{$376\times 10^{3}$} &\textcolor{red}{\bm{$96.60${\scriptsize $\pm0.01$}}} & \textcolor{red}{\bm{$95.70${\scriptsize $\pm0.02$}}}& \bm{$97.47${\scriptsize $\pm0.05$}}& \bm{$97.10${\scriptsize $\pm0.09$}}& \textcolor{violet}{\bm{$82.35${\scriptsize $\pm0.06$}}} & \textcolor{violet}{\bm{$83.67${\scriptsize $\pm0.06$}}}\\ 
    \bottomrule\\[-2.5mm]
    \multicolumn{9}{l}{$\dagger$ The top three are highlighted by \textbf{\textcolor{red}{First}}, \textbf{\textcolor{violet}{Second}}, \textbf{Third}.}
    \end{tabular}
    }
    \end{center}
\end{table*}

\begin{table}[t]
    \caption{Comparison of computational complexity and training speed.}
    \label{tab:complexity}
    \begin{center}
    \resizebox{\linewidth}{!}{
    \begin{tabular}{lrrr}
    \toprule
    \textbf{Model} & \textbf{Wikipedia} & \textbf{Reddit} & \textbf{MOOC}\\ \midrule
    \textsc{DyRep}\citep{trivedi2019dyrep} & $20.1$s {\scriptsize $\pm0.6$s} & $139.3$s {\scriptsize $\pm0.1$s} &   $78.34$s {\scriptsize $\pm0.6$s} \\
    \textsc{JODIE-rnn}\citep{kumar2019predicting} & $17.4$s {\scriptsize $\pm2.0$s} & \bm{$121.8$}s{\scriptsize $\pm0.3$s} & $62.64$s {\scriptsize $\pm0.1$s} \\ 
    \textsc{JODIE-gru}\citep{kumar2019predicting} & \bm{$16.9$}s {\scriptsize $\pm1.1$s} & $131.6$s {\scriptsize $\pm 1.5$s} & \bm{$58.82$}s {\scriptsize $\pm 2.2$s} \\
    \textsc{TGN-gru} \citep{rossi2020temporal} & $24.9$s {\scriptsize $\pm 0.3$s} & $128.1$s {\scriptsize $\pm2.2$s} & $78.11$s {\scriptsize $\pm 0.7$s}  \\
    \textsc{SWINIT-mlp} (ours) & \textcolor{red}{\bm{$9.87$}s {\scriptsize $\pm0.1$s}} & \textcolor{red}{\bm{$63.3$}s {\scriptsize $\pm1.1$s}} & \textcolor{red}{\bm{$38.41$}s {\scriptsize $\pm0.5$s}} \\ 
    \textsc{SWINIT-gru} (ours) &\textcolor{violet}{\bm{$12.5$}s {\scriptsize $\pm0.3$s}} & \textcolor{violet}{\bm{$83.6$}s {\scriptsize $\pm0.1$s}} & \textcolor{violet}{\bm{$49.20$}s {\scriptsize $\pm0.1$s}} \\ 
    \bottomrule\\[-2.5mm]
    \end{tabular}
    }
    \end{center}
\end{table}

\begin{table}[t]
    \caption{ROC-AUC of node classification over 10 repetitions.}
    \label{tab:nodeClassification}
    \begin{center}
    \resizebox{0.95\linewidth}{!}{
    \begin{tabular}{lrrr}
    \toprule
    \textbf{Model} & \textbf{Wikipedia} & \textbf{Reddit} & \textbf{MOOC} \\ \midrule
    \textsc{DyRep}\citep{trivedi2019dyrep} & $84.59${\scriptsize $\pm2.21$}&$62.91${\scriptsize $\pm2.40$} &  $69.86${\scriptsize $\pm0.02$} \\
    \textsc{JODIE-rnn}\citep{kumar2019predicting} & $85.38${\scriptsize $\pm0.08$}&$61.68${\scriptsize $\pm0.01$} &$66.82${\scriptsize $\pm0.05$} \\ 
    \textsc{JODIE-gru}\citep{kumar2019predicting} &$87.90${\scriptsize $\pm0.09$} & \bm{$64.30${\scriptsize $\pm0.21$}}& \bm{$70.23${\scriptsize $\pm0.09$}}\\
    \textsc{TGN-gru} \citep{rossi2020temporal} &\textcolor{violet}{\bm{$88.95${\scriptsize $\pm0.07$}}}& $61.49${\scriptsize $\pm0.01$} & \textcolor{violet}{\bm{$70.32${\scriptsize $\pm0.13$}}} \\
    \textsc{SWINIT-mlp} (ours) &\bm{$88.37${\scriptsize $\pm0.03$}} &\textcolor{violet}{\bm{$64.94${\scriptsize $\pm0.07$}}} & $69.52${\scriptsize $\pm0.08$} \\ 
    \textsc{SWINIT-gru} (ours) &\textcolor{red}{\bm{$90.32${\scriptsize $\pm0.05$}}}  & \textcolor{red}{\bm{$65.28${\scriptsize $\pm0.05$}}}& \textcolor{red}{\bm{$71.08${\scriptsize $\pm0.02$}}}\\ 
    \bottomrule\\[-2.5mm]
    \end{tabular}
    }
    \end{center}
\end{table}

This section reports the performance of \textsc{SWINIT} in comparison to three baseline models and five ablation studies. The main experiment tests on two link prediction tasks with both inductive and transductive learning tasks. The best reported performance are tuned with PyTorch on NVIDIA\textsuperscript{\textregistered} Tesla V100 GPU with 5,120 CUDA cores and 16GB HBM2 mounted on an HPC cluster. Experimental code in PyTorch can be found at \url{https://github.com/bzho3923/GNN\_SWINIT}.

\subsection{Experimental Protocol}
\subsubsection{Dataset}
Our experiments are conducted on three bipartite graph datasets: \textbf{Wikipedia}, \textbf{Reddit} and \textbf{MOOC} \citep{kumar2019predicting,liyanagunawardena2013moocs}. 
\begin{itemize}
    \item \textbf{Wikipedia} has users and Wikipedia page as the two sets of nodes. An edge is recorded when a user edits a page. The dataset selects the $1,000$ most edited pages and frequent editing users who made at least $5$ edits. The dataset contains $9,227$ nodes and $157,474$ edges in total, and each event is described by $172$ features. 
    \item \textbf{Reddit} divides two sets of nodes as users and subreddits (communities). An interaction occurs when a user posts a message to a subreddit. The datasets samples $1,000$ most active subreddits as nodes along with the $10,000$ most active users. In total, the dataset contains $11,000$ nodes and $672,447$ edges. All events are recorded as $172$ edge features by the LIWC categories \citep{pennebaker2001linguistic} by the text of each post.
    \item \textbf{MOOC} records students and courses of the ``Massive Open Online Course" learning platform. An interaction occurs when a student enrolls in the course. The dataset consists of $7,047$ students, $97$ courses and $411,749$ interactions. Specifically, $4,066$ state changes are recorded implying action that a student drops out of a course. 
\end{itemize}

\subsubsection{Learning Tasks}
We conduct prediction tasks of the three datasets on link prediction and node classification. 

In the link prediction tasks, the goal is to predict the probability of an edge occurring between two nodes at a given time. To conduct the prediction, a simple MLP decoder is concatenated after the node embedding module, mapping from the concatenation of the node pair's embeddings to the probability of the edge. Unlike tabular data where entities are independent of each other, graph nodes are inter-connected. Different graph data splitting strategies can affect the performance of prediction tasks. Depending on the degree of information accessed in the test set, we split our dynamic predicting tasks into inductive and transductive settings. \textbf{Transductive} setting refers to the reasoning from the observed graph to specific nodes or edges \citep{joachims2003transductive}. It assumes that the entire input graph can be observed across the training, validation, and test sets. Our predictions for the future edges are based on the observed input graph. \textbf{Inductive} setting refers to the reasoning from observed graph information to general rules (or unseen graphs, in our case) \citep{hamilton2017inductive}. We break edges to get independent training, validation, and test graphs. The task is to predict future links of unseen nodes. 

In node classification, the target is to predict whether the linkage between users and items will lead to a state change in users \cite{kumar2016ask}. In particular, a trained model tries to predict if a user will be banned (Wikipedia, Reddit) or if a student will drop out from the course (MOOC). Note that the three datasets are highly-imbalanced and the positive rate is at most $1\%$, since the chances a user will be banned or a student' drop-out should be small. In this task, a trained model from the link prediction task serves as the encoder. A trainable MLP decoder mapping from the node embedding to the probability of state change.

\subsubsection{Evaluation Metrics}
We follow the design of prediction task by PyTorch on interactions happening in time $t+1$ and $t+2$ of a bipartite graph given information until time $t$. The evaluation of our model is based on two specific metrics: \textbf{precision} and \textbf{ROC-AUC}. 
\begin{itemize}
    \item \textbf{Precision} refers to the number of true positives (TP) divided by the total number of positive observations, i.e., the sum of true positive and false positive (TP$+$FP) instances. The information retrieval theory \citep{baeza1999modern} suggests the best model has its precision at $1$, which implies that labels of all the positive samples are correctly predicted. In our cases, we use precision to measure the proportion of predicted interactions that indeed exists in the ground truth dataset.
    \item \textbf{ROC-AUC}, or `The Area under the ROC Curve' where ROC stands for receiver operating characteristic curve is a standard performance measure for classification tasks \citep{bradley1997use}. ROC-AUC calculates the 2D area underneath the ROC curve. It measures models' capability of distinguishing labels among classes. Here we use ROC-AUC to measure the probability that the model ranks a random true interaction that is higher than a random false interaction. 
\end{itemize}

\subsubsection{Comparison Baselines}
\textsc{SWINIT} is compared with three dynamic graph models for continuous-time inputs. \textsc{JODIE}  \citep{kumar2019predicting} \footnote{\url{https://github.com/srijankr/jodie}}  employs two recurrent neural network models and introduces an innovative projection function that learns the future embeddings of any user. \textsc{DyRep} \citep{trivedi2019dyrep} \footnote{implemented by \url{https://github.com/twitter-research/tgn}} designs a latent mediation process to capture the topological evolution and node-level interactions. The node neighbours are aggregated by \textsc{GAT} \citep{velivckovic2017graph}.
\textsc{TGN} \citep{rossi2020temporal} \footnote{\url{https://github.com/twitter-research/tgn}} develops a novel memory module to consider the long-term dependencies in dynamic graphs and employs temporal graph attention to learn the temporal embeddings.

\subsubsection{Training Setup}
We follow the pseudo-code of Algorithm~\ref{alg:SWINIT} and design \textsc{SWINIT} accordingly. In the spectral attention module, we approximate truncated SVD with some largest modes with $q$-iteration. The specific number of node is selected as the smallest number between $50$ and $100$ such that the spectral norm error is less than $0.1$. The batch memory window is processed by fully connected layers, and the prepared subgraphs are then processed by \textsc{UFGConv} with Haar-type filters at dilation factor $2^l$ to allow efficient transforms. To train a generalized model that is robust to small disturbance, in the validation set we randomly add 50\% negative samples at each epoch. The same negative sampling procedure is conducted in the test set, except that all the samples are deterministic. This design is universally used in literature so that the prediction tasks become non-trivial. The hyper-parameters of baseline models, unless specified, are fixed to the best choice provided by their authors. For all the models, we fix the batch size at $1,000$ with a maximum of $200$ epochs for both datasets. Any employed neural network overlays either $2$ or $3$ layers, and the memory dimension, node embedding dimension, time embedding dimension are selected from $\{100,150,200\}$ respectively. To make the comparison as fair as possible, the number of parameters of each model corresponding the fine tuned hyperparameters are reflected in Table \ref{tab:linkPrediction}.  The optimal learning rate is tuned from the range of $\{1\mathrm{e}-4,5\mathrm{e}-5\}$, and the weight decay is fixed at $1\mathrm{e}-2$. The training process is optimized by \textsc{AdamW} \citep{loshchilov2018decoupled}.  All the datasets follow the standard split and processing rules as in \citep{kumar2019predicting,rossi2020temporal}. The average test accuracy and its standard deviation come from $10$ runs. 


\begin{table}[t]
    \caption{Average performance for ABLATION study on Wikipedia.}
    \label{tab:ablation_wiki}
    \begin{center}
    \resizebox{\linewidth}{!}{
    \begin{tabular}{lrrr}
    \toprule
    & \multicolumn{2}{c}{\textbf{link prediction}} & \textbf{node classification} \\  \cmidrule(lr){2-3}\cmidrule(lr){4-4}
    \textbf{Module} & precision & ROC-AUC & ROC-AUC \\ \midrule
    \textsc{raw+mlp} & $96.71${\scriptsize $\pm0.10$} & $96.20${\scriptsize $\pm0.20$} & $88.33${\scriptsize $\pm0.18$} \\ 
    \textsc{raw+trSVD+mlp} & $96.82${\scriptsize $\pm0.13$} & $96.23${\scriptsize $\pm0.12$} & \bm{$88.47${\scriptsize $\pm0.22$}} \\ 
    \textsc{trSVD+mlp} & \bm{$97.02${\scriptsize $\pm0.06$}} & \bm{$96.51${\scriptsize $\pm0.08$}} & $88.37${\scriptsize $\pm0.03$} \\ \midrule
    \textsc{raw+gru} & $97.18${\scriptsize $\pm0.03$} & $96.84${\scriptsize $\pm0.04$} & $88.66${\scriptsize $\pm0.27$} \\ 
    \textsc{raw+trSVD+gru} & $97.23${\scriptsize $\pm0.04$} & $96.90${\scriptsize $\pm0.05$} & $87.56${\scriptsize $\pm0.09$} \\ 
    \textsc{trSVD+gru} & \bm{$97.44${\scriptsize $\pm0.05$}} & \bm{$97.15${\scriptsize $\pm0.06$}} & \bm{$90.32${\scriptsize $\pm0.05$}} \\
    \bottomrule\\[-2.5mm]
    \end{tabular}
    }
    \end{center}
\end{table}


\subsection{Result Analysis with Baseline Comparison}
\subsubsection{Prediction Performance}
We report the performance of transductive and inductive link prediction tasks in Table~\ref{tab:linkPrediction}. In addition, we provide the average computation speed per epoch and the number of parameters to be estimated in Table~\ref{tab:complexity}. Of all four scenarios (two tasks on two settings), \textsc{SWINIT} outperforms \textsc{JODIE} and \textsc{DyRep}, and achieves at least comparable performance to \textsc{TGN} for both precision and ROC-AUC metric. The standard deviation is also controlled at a low level, which reflects our model has consistent performance over time. It is worth noticing that \textsc{JODIE-gru} outperforms the original \textsc{JODIE-rnn} adopted by the authors of \cite{kumar2019predicting}. The performance gain of GRU over RNN explains to some extent the outperformance of \textsc{TGN} over \textsc{SWINIT-mlp} in some cases. We would like to stress here that our memory unit is propagated by MLP, which is an even simpler design compared to any recurrent unit. While the disadvantage is considerably small (or even twisted), it stresses the effectiveness of the spectral modules we rely on. This observation can easily be verified by the results of \textsc{SWINIT-gru} where the memory layer adopts the GRU module. It outperforms all other baselines significantly, including \textsc{TGN}, over different datasets, learning tasks, and evaluation metrics. 

The average AOC-ROC score for node classification tasks on the three datasets are reported in Table~\ref{tab:nodeClassification}. As mentioned in experimental setup, the extremely imbalanced nature of node classes results in lower performance of all the models on average. Still, \textsc{SWINIT} outperforms the baselines, where the advantage is greater when equipping GRU module. In particular, GRU-enhanced models achieve the highest ROC-AUC scores of all time.

\subsubsection{Computational Complexity}
In supplement to the performance of prediction tasks, we provide the average computation speed per epoch and the number of parameters to be estimated in Table~\ref{tab:complexity}. 
In general, the computational complexity, \textsc{SWINIT} requires \textbf{SEVEN} times smaller than \textsc{TGN-gru} the amount of parameters to approximate, in comparison to \textsc{TGN-gru}. The computational speed of \textsc{SWINIT} is mainly influenced by the frequent operation of pushing and pulling $\theta$s from the global $\Theta$. This issue becomes obvious when the total number of edges in the dataset grows to a considerably large mass. With further optimization on a practice level, we believe the speed issue can be well-resolved or at least lessen to an acceptable level. We leave the improvements for future work. Despite the cost of \textsc{UFGConv}, \textsc{SWINIT} propagates faster than all other baselines. Particularly on the Reddit dataset, which has more than $4$ times of events compared to Wikipedia, \textsc{SWINIT} boost at least $50\%$ speed while maintaining top performance. This result is consistent with the analysis in Section~\ref{sec:attnSVD} and Section~\ref{sec:model}, where the computational cost of \textsc{SWINIT} is significantly reduced through calculating $\mX^{\top}\mX$ in priority. 


\begin{remark}
    The performance of \textsc{TGN} in both transductive and inductive settings are lower than scores reported in \cite{rossi2020temporal}. The difference lies in the access of previous interactions. In particular, the task we (also in PYG) design is constrained on a more realistic scenario that no ground truth after $t$ can be accessed when we validate and test at time $t$ and $t+1$. The prediction is thus made in parallel. In contrast, \textsc{TGN} has access to all previous data when sampling node neighborhoods for interactions later in the batch.
\end{remark}

\subsection{Ablation Study}
In addition to the comparison against baseline methods, we also justify the choice of the temporal encoder, including the spectral attention and sequential network. For the spectral attention module, we set up three different data encoders, the input of which is the raw data, and the output is a transformed data matrix that is used for batch training (graph slicing). The three encoders differ in the proportion of used column basis from SVD: no SVD transform (\textsc{raw}), truncated SVD transform that concatenates raw input (\textsc{raw + trSVD}), and truncated SVD transform (\textsc{trSVD}). The last model, i.e., \textsc{trSVD}, is used as our final model. For the sequential temporal information encoder, we consider MLP and GRU modules, which are also the two choices we evaluated in the last two experiments. A total number of six models are thus to be validated. Note that when comparing different choices of attention, we exclude the vanilla transformer encoder, as its speed is too slow to provide the average score within a limited computing resource. As a matter of fact, an epoch of the self-attention encoder requires more than $30$ minutes to run. In contrast, it only takes \textsc{trSVD} less than $1.5$ minutes to finish an epoch on a model of the same setting. 

The models are validated on Wikipedia of both link prediction and node classification tasks. We follow the same setups aligned with Table~\ref{tab:linkPrediction}. The hyper-parameters are fixed to the optimal results from the best performed \textsc{SWINIT} in the earlier baseline comparison experiment. As the prediction performance reported in Table~\ref{tab:ablation_wiki}, all the three variants of \textsc{SWINIT} achieve a similar level of outstanding scores, which is due to the main architectural design. Still, \textsc{trSVD} outperforms \textsc{raw + trSVD} with a noticeable differences, and achieves a slightly better performance than \textsc{raw}. The outperformance is much more significant with GRU than MLP network, where the former contains more complicated architectures for local temporal information aggregation. To understand such observation, a possible interpretation is that the importance of feature abbreviation and pattern extraction gains increasing importance when the feature-to-sample ratio increases. The smaller the sample size is, the more harmful a redundant data input becomes. The observations demonstrate the effectiveness of our design regarding encoding principal patterns of both dimensions.

\section{Discussion}
\label{sec:conclusion}
The paper proposed a new graph neural network framework for the prediction tasks of temporal data. The operations in the computational unit are fully defined by a trainable spectral transform, where the trainable neural unit is implemented by fully connected layers, i.e., MLP. It connects the two spectral-based modules. The first module uses the randomized SVD for all the event edges of the dynamic graph. It extracts the key temporal node features. By transforming features to the spectral domain, the node dimension and time dimension are fused properly. With SVD, the conventional multi-layer perception becomes as powerful as complex recurrent memory modules, such as GRU and LSTM, in dynamic graph prediction tasks. MLP takes the feature batches of subgraphs from SVD and outputs the trained features to the framelet convolution (\textsc{UFGConv}). The latter has been proved able to capture the graph structure information in a multi-level and multi-scale learning presentation \cite{zheng2021framelets}.  The graph structure is mainly embedded in the framelet basis of the framelet convolution. The experiments show that the framelet system constructed by the original data has a strong transferrability in the dynamic evolution of graph data. Thus, the three units, SVD, MLP, and \textsc{UFGConv}, altogether serve as a powerful network engine for Dynamic GNNs to learn time-dependent structured data. Both theoretical and empirical evidence demonstrates that the learnable SVD equipped with randomized SVD and MLP plays an equal role as a transformer. Inherited from the scalability of the SVD and especially the framelet transforms-based \textsc{UFGConv}, \textsc{SWINIT} then achieves better scalability than traditional transformer encoders.



\bibliographystyle{ACM-Reference-Format}
\bibliography{reference}

\end{document}

%% file: math_commands.tex
\usepackage{amsmath,amsfonts,bm}

\def\eqref#1{equation~\ref{#1}}

\def\1{\bm{1}}

\def\vtheta{{\bm{\theta}}}

\def\vu{{\bm{u}}}

\def\vx{{\bm{x}}}

\def\mA{{\bm{A}}}

\def\mH{{\bm{H}}}

\def\mK{{\bm{K}}}

\def\mQ{{\bm{Q}}}
\def\mR{{\bm{R}}}

\def\mU{{\bm{U}}}
\def\mV{{\bm{V}}}
\def\mW{{\bm{W}}}
\def\mX{{\bm{X}}}
\def\mY{{\bm{Y}}}

\DeclareMathAlphabet{\mathsfit}{\encodingdefault}{\sfdefault}{m}{sl}
\SetMathAlphabet{\mathsfit}{bold}{\encodingdefault}{\sfdefault}{bx}{n}

\def\gG{{\mathcal{G}}}

\def\gL{{\mathcal{L}}}

\def\gO{{\mathcal{O}}}

\def\gT{{\mathcal{T}}}

\def\gV{{\mathcal{V}}}
\def\gW{{\mathcal{W}}}

\def\sV{{\mathbb{V}}}

\newcommand{\R}{\mathbb{R}}



%% file: tikz_SVD.tex
\newcommand\initialy{4}
\newcommand\nodeSize{0.75cm}

\tikzset{%
  tipSquare/.tip={Circle[open]}
}

\tikzset{%
  every neuron/.style={
    circle,
    draw,
    fill=white,
    scale=0.8,
    minimum size=\nodeSize
  },
  neuron missing/.style={
    draw=none, 
    scale=0.8,
    fill=$\dots$,
    text height=0cm,
    execute at begin node=$\dots$
  },
  snake it/.style={
    decorate, decoration=snake
  }
}

\begin{tikzpicture}[x=1.5cm, y=1.5cm, >=stealth, scale=0.58, every node/.style={transform shape}, curved arrow/.style={arc arrow={to pos #1 with length 2mm and options {}}},
reversed curved arrow/.style={arc arrow={to pos #1 with length 2mm and options reversed}}]  

\def\xstart{-3.4}
\def\ystart{-1}
\draw[rounded corners=0mm, fill=black!10!orange!70!, opacity=0.6] (\xstart,\ystart) --++ (1,0) --++ (0,-2) --++ (-1,0) -- cycle;

\node[align=center, above] at (\xstart+1.2, \ystart-0.3) {$\approx$};
\node[align=center, above] at (\xstart+0.5, \ystart-0.3) {$\widetilde{\mX}$};
\node[align=center, above] at (\xstart+0.5, \ystart) {$N\times d'$};

\draw[rounded corners=0mm, fill=black!10!orange!70!] (\xstart,\ystart-1) --++ (1,0) --++ (0,-0.15) --++ (-1,0) -- cycle;
\node[every neuron, fill=black!10!red!50!, minimum size=0.2cm] at (\xstart+0.25,\ystart-1-0.07){};

\node[align=center, above] at (\xstart+0.25, \ystart-1.55) {$\widetilde{\mX}_{ij}$};

\def\xstart{-2}
\def\ystart{-1}
\draw[rounded corners=0mm, fill=black!15!yellow!40!, opacity=0.6] (\xstart,\ystart) --++ (1,0) --++ (0,-2) --++ (-1,0) -- cycle;
\node[align=center, above] at (\xstart+1.125, \ystart-0.3) {$\times$};
\draw[rounded corners=0mm, fill=black!20!green!30!, opacity=0.6] (\xstart+1.25,\ystart) --++ (0.8,0) --++ (0,-1) --++ (-0.8,0) -- cycle;

\node[align=center, above] at (\xstart+2.4, \ystart-0.3) {$\approx$};
\node[align=center, above] at (\xstart+0.5, \ystart-0.3) {$\mX$};
\node[align=center, above] at (\xstart+1.25+0.7, \ystart-0.3) {$\mV$};
\node[align=center, above] at (\xstart+0.5, \ystart) {$N\times d$};
\node[align=center, above] at (\xstart+1.25+0.5, \ystart) {$d\times d'$};

\draw[rounded corners=0mm, fill=black!15!yellow!50!] (\xstart,\ystart-1) --++ (1,0) --++ (0,-0.15) --++ (-1,0) -- cycle;
\draw[rounded corners=0mm, fill=black!20!green!40!] (\xstart+1.25+0.3,\ystart) --++ (0.15,0) --++ (0,-1) --++ (-0.15,0) -- cycle;

\node[align=center, above] at (\xstart+0.5, \ystart-1.4) {$i$th row};
\node[align=center, above] at (\xstart+1.25+0.5, \ystart-1.3) {$j$th column};

\def\xstart{0.55}
\def\ystart{-1}

\draw[rounded corners=0.5mm, outer sep=0pt, fill= black!20!green!30!, opacity=0.1] (\xstart+1.2,\ystart+0.1) --++ (4.5,0) --++ (0,-2.8) --++ (-4.5,0) -- cycle;
\draw[rounded corners=0.5mm, dashed, black!35!green!55!] (\xstart+1.2,\ystart+0.1) --++ (4.5,0) --++ (0,-2.8) --++ (-4.5,0) -- cycle;

\draw[rounded corners=0mm, fill=black!15!yellow!40!, opacity=0.6] (\xstart,\ystart) --++ (1,0) --++ (0,-2) --++ (-1,0) -- cycle;

\draw[rounded corners=0mm, fill=black!15!yellow!40!, opacity=0.6] (\xstart+1.3,\ystart) --++ (2,0) --++ (0,-1) --++ (-2,0) -- cycle;
\draw[rounded corners=0mm, fill=black!15!yellow!40!, opacity=0.6] (\xstart+3.45,\ystart) --++ (1,0) --++ (0,-2) --++ (-1,0) -- cycle;
\draw[rounded corners=0mm, fill=black!20!cyan!50!, opacity=0.6] (\xstart+4.6,\ystart) --++ (0.8,0) --++ (0,-1) --++ (-0.8,0) -- cycle;

%
\node[align=center, above] at (\xstart+0.5, \ystart-0.3) {$\mX$};
\node[align=center, above] at (\xstart+1.3+1, \ystart-0.3) {$\mX^{\top}$};
\node[align=center, above] at (\xstart+3.45+0.7, \ystart-0.3) {$\mX$};
\node[align=center, above] at (\xstart+4.6+0.75, \ystart-0.3) {$\mathbf{\Omega}\mR^{-1}$};

\node[align=center, above] at (\xstart+0.5, \ystart) {$N\times d$};
\node[align=center, above] at (\xstart+1.3+1, \ystart) {$d\times N$};
\node[align=center, above] at (\xstart+3.45+0.5, \ystart) {$N\times d$};
\node[align=center, above] at (\xstart+4.6+0.5, \ystart) {$d\times d'$};

\draw[rounded corners=0mm, fill=black!15!yellow!50!] (\xstart,\ystart-1) --++ (1,0) --++ (0,-0.15) --++ (-1,0) -- cycle;
\draw[rounded corners=0mm, fill=black!15!yellow!50!] (\xstart+1.3,\ystart-0.3) --++ (2,0) --++ (0,-0.15) --++ (-2,0) -- cycle;
\draw[rounded corners=0mm, fill=black!15!yellow!50!] (\xstart+3.45+0.3,\ystart) --++ (0.15,0) --++ (0,-2) --++ (-0.15,0) -- cycle;
\draw[rounded corners=0mm, fill=black!20!cyan!60!] (\xstart+4.6+0.3,\ystart) --++ (0.15,0) --++ (0,-1) --++ (-0.15,0) -- cycle;

\draw[rounded corners=0mm, fill=black!15!yellow!60!, opacity=0.6] (\xstart+2,\ystart-1.5) --++ (1,0) --++ (0,-1) --++ (-1,0) -- cycle;
\draw[rounded corners=0mm, fill=black!15!yellow!60!] (\xstart+2,\ystart-1.5-0.25) --++ (1,0) --++ (0,-0.15) --++ (-1,0) -- cycle;
\draw[rounded corners=0mm, fill=black!35!yellow!60!] (\xstart+2,\ystart-1.5-0.45) --++ (1,0) --++ (0,-0.15) --++ (-1,0) -- cycle;
\draw[rounded corners=0mm, fill=black!55!yellow!60!] (\xstart+2,\ystart-1.5-0.65) --++ (1,0) --++ (0,-0.15) --++ (-1,0) -- cycle;
\node[every neuron, fill=black!10!orange!50!, minimum size=0.2cm] at (\xstart+2+0.3+0.07,\ystart-1.5-0.24-0.07){};

\node[align=center, above] at (\xstart+0.5, \ystart-1.4) {$i$th row};
\node[align=center, above] at (\xstart+1.3+1, \ystart-0.75) {$j$th row};
\node[align=center, above] at (\xstart+3.45+0.5, \ystart-2.4) {$j$th column};
\node[align=center, above] at (\xstart+4.6+0.5, \ystart-1.3) {$j$th column};

\draw[rounded corners,densely dotted,line width=0.2mm,black!50!] [<-] (-3.1,-2.5) --++ (0,-1) --++ (2.8,0) --++ (0,1.2);
\draw[rounded corners,densely dotted,line width=0.2mm,black!50!] (-1.3,-2.5) --++ (0,-0.9);
\node[every neuron,minimum size=0.4cm] at (-1.3,-2.5-1) {$\times$};
\node[align=center, above] at (-2.2,-2.5-1.05) {interpretation:1};

\draw[rounded corners,densely dotted,line width=0.2mm,black!50!][->] (2.9,-1.7) --++ (0,-1);
\draw[rounded corners,densely dotted,line width=0.2mm,black!50!] (4.2,-2.2) --++ (-1.3,0);
\draw[rounded corners,densely dotted,line width=0.2mm,black!50!] (3.2,-4) --++ (0,0.5);

\draw[rounded corners,densely dotted,line width=0.2mm,black!50!] [<-] (-3.3,-2.5) --++ (0,-1.5) --++ (8.8,0) --++ (0,1.7);
\draw[rounded corners,densely dotted,line width=0.2mm,black!50!] (1,-2.5) --++ (0,-1.5);

\node[every neuron,minimum size=0.4cm] at (2.9,-2.2) {$\times$};
\node[every neuron,minimum size=0.4cm] at (3.2,-2.5-1.5) {$\times$};
\node[every neuron,minimum size=0.4cm] at (1,-2.5-1.5) {$\times$};
\node[align=center, above] at (-2.3,-2.5-1.55) {interpretation:2};

\end{tikzpicture}

%% file: tikz_UFG.tex
\newcommand\initialy{4}
\newcommand\nodeSize{0.75cm}

\tikzset{%
  tipSquare/.tip={Circle[open]}
}

\tikzset{%
  every neuron/.style={
    circle,
    draw,
    fill=white,
    scale=0.8,
    minimum size=\nodeSize
  },
  neuron missing/.style={
    draw=none, 
    scale=0.8,
    fill=$\dots$,
    text height=0cm,
    execute at begin node=$\dots$
  },
  snake it/.style={
    decorate, decoration=snake
  }
}

\begin{tikzpicture}[x=1.5cm, y=1.5cm, >=stealth, scale=0.58, every node/.style={transform shape}, curved arrow/.style={arc arrow={to pos #1 with length 2mm and options {}}},
reversed curved arrow/.style={arc arrow={to pos #1 with length 2mm and options reversed}}]  

\draw[->,dashed](-2,-1.0)--++(0.3,0);
\draw[->,dashed](2.6,-1.0)--++(0.3,0);

\draw[rounded corners=0mm, dashed, black!30!yellow!50!, fill=black!5!yellow!15!, opacity=0.6] (-4,-0.17) -- (-3.4,-0.83) --++ (1.2,0) -- (-2.8,-0.17) -- cycle;
\draw[rounded corners=0mm, dashed, black!30!yellow!50!, fill=black!5!yellow!15!, opacity=0.6] (-0.5,-0.17) -- (-1.58,-0.83) --++ (1.2,0) -- (0.7,-0.17) -- cycle;

\draw[rounded corners=0mm, dashed, black!40!green!40!, fill=black!30!green!10!, opacity=0.6] (-0.1,-0.17) -- (0.8,-0.83) --++ (1.6,0) -- (1.5,-0.17) -- cycle;
\draw[rounded corners=0mm, dashed, black!40!green!40!, fill=black!30!green!10!, opacity=0.6] (3.6,-0.17) -- (3.0,-0.83) --++ (1.6,0) -- (5.2,-0.17) -- cycle;

\draw[rounded corners=0mm, dashed, black!30!yellow!50!] (-4,-0.17) -- (-3.4,-0.83) --++ (1.2,0) -- (-2.8,-0.17) -- cycle;
\draw[rounded corners=0mm, dashed, black!30!yellow!50!] (-0.5,-0.17) -- (-1.58,-0.83) --++ (1.2,0) -- (0.7,-0.17) -- cycle;

\draw[rounded corners=0mm, dashed, black!40!green!40!] (-0.1,-0.17) -- (0.8,-0.83) --++ (1.6,0) -- (1.5,-0.17) -- cycle;
\draw[rounded corners=0mm, dashed, black!40!green!40!] (3.6,-0.17) -- (3.0,-0.83) --++ (1.6,0) -- (5.2,-0.17) -- cycle;

\draw[->,line width=0.3mm,dashed,black!30!yellow!50!](-3.3,-0.3)--(-2.9,-0.7);
\draw[->,line width=0.3mm,dashed,black!30!yellow!50!](-0.8,-0.7)--(0,-0.3);
\draw[->,line width=0.3mm,dashed,black!40!green!40!](0.9,-0.3)--(1.4,-0.7);
\draw[->,line width=0.3mm,dashed,black!40!green!40!](3.9,-0.7)--(4.3,-0.3);

\def\ystart{0}
\def\xstart{-3.8}
\node[draw, align=center, fill=black!10!] at (\xstart, \ystart){$\theta_1$};
\node[draw, align=center, fill=black!10!] at (\xstart+0.4, \ystart){$\theta_2$};
\node[draw, align=center, fill=black!10!] at (\xstart+0.8, \ystart){$\theta_3$};
\node[draw, align=center, fill=black!10!] at (\xstart+1.2, \ystart){$\theta_4$};
\node[draw, align=center, fill=black!10!] at (\xstart+1.6, \ystart){$\theta_5$};

\def\xstart{-0.3}
\node[draw, align=center, fill=black!20!yellow!50!] at (\xstart, \ystart){$\theta_1$};
\node[draw, align=center, fill=black!20!yellow!50!] at (\xstart+0.4, \ystart){$\theta_2$};
\node[draw, align=center, fill=black!20!yellow!50!] at (\xstart+0.8, \ystart){$\theta_3$};
\node[draw, align=center, fill=black!10!] at (\xstart+1.2, \ystart){$\theta_4$};
\node[draw, align=center, fill=black!10!] at (\xstart+1.6, \ystart){$\theta_5$};

\def\xstart{3.4}
\node[draw, align=center, fill=black!20!yellow!50!] at (\xstart, \ystart){$\theta_1$};
\node[draw, align=center, fill=black!30!green!30!] at (\xstart+0.4, \ystart){$\theta_2$};
\node[draw, align=center, fill=black!30!green!30!] at (\xstart+0.8, \ystart){$\theta_3$};
\node[draw, align=center, fill=black!30!green!30!] at (\xstart+1.2, \ystart){$\theta_4$};
\node[draw, align=center, fill=black!30!green!30!] at (\xstart+1.6, \ystart){$\theta_5$};

\def\ystart{-1.0}
\def\xstart{-3.2}
\node[draw, align=center, fill=black!10!] at (\xstart, \ystart){$\theta_1$};
\node[draw, align=center, fill=black!10!] at (\xstart+0.4, \ystart){$\theta_2$};
\node[draw, align=center, fill=black!10!] at (\xstart+0.8, \ystart){$\theta_3$};

\def\xstart{-1.4}
\node[draw, align=center, fill=black!20!yellow!50!] at (\xstart, \ystart){$\theta_1$};
\node[draw, align=center, fill=black!20!yellow!50!] at (\xstart+0.4, \ystart){$\theta_2$};
\node[draw, align=center, fill=black!20!yellow!50!] at (\xstart+0.8, \ystart){$\theta_3$};

\def\xstart{0.6}
\node[draw, align=center, fill=black!20!yellow!50!] at (\xstart+0.4, \ystart){$\theta_2$};
\node[draw, align=center, fill=black!20!yellow!50!] at (\xstart+0.8, \ystart){$\theta_3$};
\node[draw, align=center, fill=black!10!] at (\xstart+1.2, \ystart){$\theta_4$};
\node[draw, align=center, fill=black!10!] at (\xstart+1.6, \ystart){$\theta_5$};

\def\xstart{2.8}
\node[draw, align=center, fill=black!40!green!30!] at (\xstart+0.4, \ystart){$\theta_2$};
\node[draw, align=center, fill=black!40!green!30!] at (\xstart+0.8, \ystart){$\theta_3$};
\node[draw, align=center, fill=black!40!green!30!] at (\xstart+1.2, \ystart){$\theta_4$};
\node[draw, align=center, fill=black!40!green!30!] at (\xstart+1.6, \ystart){$\theta_5$};

\node[align=center, above] at (-3.0, 0.2) {$\mathbf{\Theta}[t-1]$};
\node[align=center, above] at (0.5, 0.2) {$\mathbf{\Theta}[t]$};
\node[align=center, above] at (4.2, 0.2) {$\mathbf{\Theta}[t+1]$};

\node[align=center, above] at (-2.8, -1.55) {$\theta^{(0)}[t]$};
\node[align=center, above] at (-1.0, -1.55) {$\theta^{(n)}[t]$};
\node[align=center, above] at (1.6, -1.55) {$\theta^{(0)}[t+1]$};
\node[align=center, above] at (3.8, -1.55) {$\theta^{(n)}[t+1]$};

\node[align=center, above] at (-1.8, -1.9) {\textsc{UFGConv}$[t]$};
\node[align=center, above] at (2.6, -1.9) {\textsc{UFGConv}$[t+1]$};

\end{tikzpicture}